# A Data-Centric Approach to Detecting and Mitigating Demographic Bias in Pediatric Mental Health Text: A Case Study in Anxiety Detection

**Julia Ive**[1,*], **Paulina Bondaronek**[1], **Vishal Yadav**[2], **Daniel Santel**[3], **Tracy Glauser**[4], **Tina Cheng**[7], **Jeffrey R. Strawn**[8], **Greeshma Agasthya**[4], **Jordan Tschida**[4], **Sanghyun Choo**[4], **Mayanka Chandrashekar**[4], **Anuj J. Kapadia**[4], and **John Pestian**[6]

[1]University College London, Institute of Health Informatics, UK
[2]Queen Mary University of London, School of Electronic Engineering and Computer Science, London, UK
[3]Division of Biomedical Informatics, Cincinnati Children's Hospital Medical Center, University of Cincinnati, Cincinnati, OH, USA
[4]Advanced Computing in Health Sciences, Computational Sciences in Engineering Division, Oak Ridge National Laboratory
[5]Division of Neurology, Cincinnati Children's Hospital Medical Center, University of Cincinnati, Cincinnati, OH, USA
[6]Division of Biomedical Informatics, Cincinnati Children's Hospital Medical Center, University of Cincinnati, Cincinnati, OH, USA
[7]Department of Pediatrics, Cincinnati Children's Hospital Medical Center, University of Cincinnati, Cincinnati, OH, USA
[8]Department of Psychiatry, College of Medicine, University of Cincinnati, Cincinnati, OH, USA
[*]j.ive@ucl.ac.uk

## ABSTRACT

**Introduction** Healthcare analytics and Artificial Intelligence (AI) hold transformative potential, yet AI models often inherit biases from their training data, which can exacerbate healthcare disparities, particularly among minority groups. While efforts have primarily targeted bias in structured data, mental health heavily depends on unstructured data like clinical notes, where bias and data sparsity introduce unique challenges. This study aims to detect and mitigate linguistic differences related to non-biological differences in the training data of AI models designed to assist in pediatric mental health screening.
Our objectives are: (1) to assess the presence of bias by evaluating outcome parity across sex subgroups, (2) to identify bias sources through textual distribution analysis, and (3) to develop and evaluate a de-biasing method for mental health text data.
**Methods** We examined classification parity across demographic groups, identifying biases through analysis of linguistic patterns in clinical notes. Using interpretability techniques, we assessed how gendered language influences model predictions. We then applied a data-centric de-biasing method focused on neutralizing biased terms and retaining only the salient clinical information. This methodology was tested on a model for automatic anxiety detection in pediatric patients—a crucial application given the rise in youth anxiety post-COVID-19.
**Results** Our findings show a systematic under-diagnosis of female adolescent patients, with a 4% lower accuracy and a 9% higher False Negative Rate (FNR) compared to male patients, likely due to disparities in information density and linguistic differences in patient notes. Notes for male patients were on average 500 words longer, and linguistic similarity metrics indicated distinct word distributions between genders. Implementing our de-biasing approach reduced this diagnostic bias by up to 27%, demonstrating the approach's effectiveness in enhancing equity across demographic groups
**Discussion** We developed and evaluated a data-centric de-biasing framework to address gender-based content disparities within clinical text, specifically in pediatric anxiety detection. By neutralizing biased language and enhancing focus on clinically essential information, our approach highlights an effective strategy for mitigating bias in AI healthcare models trained on unstructured data. This work emphasizes the importance of developing bias mitigation techniques tailored for healthcare text, advancing equitable AI-driven solutions in mental health.

## Introduction

The global pandemic has acted as a catalyst and highlighted the changes required in health and social care systems to ensure the ongoing well-being of the population, with an emphasis on mental health. This is especially true for children and adolescents with the prevalence of anxiety and depression symptoms doubled during the pandemic[1]. These increases, particularly in older and female adolescents, highlight an urgent need for mental health help and equitable early detection efforts to mitigate the long-term impact[2]. Even without the additional load, comprehensive screening for pediatric mental health concerns is challenging[3]. AI constitutes a promising solution for expanding mental health diagnostics[4].

However, in spite of the promise of AI to assist in mental health, the development and deployment of machine learning models into real clinical environments remains limited[4,5]. One of the potential reasons is the risk of propagating harmful biases. Additionally, clinicians often have concerns about the interpretability of predictions from "black box" models obscuring model suggestions.

One crucial aspect that influences the development and implementation of building trustworthy AI models in clinical setting is the availability of high-quality data in sufficient volumes[6]. This is particularly challenging in mental health care the primary source of information in mental health care is free text (clinical notes) containing highly sensitive information. However, the available data are often sparse and biased because their selection process reflects an underlying unfairness within the healthcare system (known as selection bias) (see Figure 1). Unfortunately, AI models are prone to exaggerating this bias in a self-reinforcing cycle: when trained on data that reflect historical inequalities, these models tend to perform worse for marginalized groups, and leading to inequitable in treatment outcomes, and amplifying existing inequalities[2,7]. A recent study by Yates Coley et al.[8] found poor performance in predicting the risk of suicide for underrepresented demographic groups, including Black, Native American, and Alaskan patients.

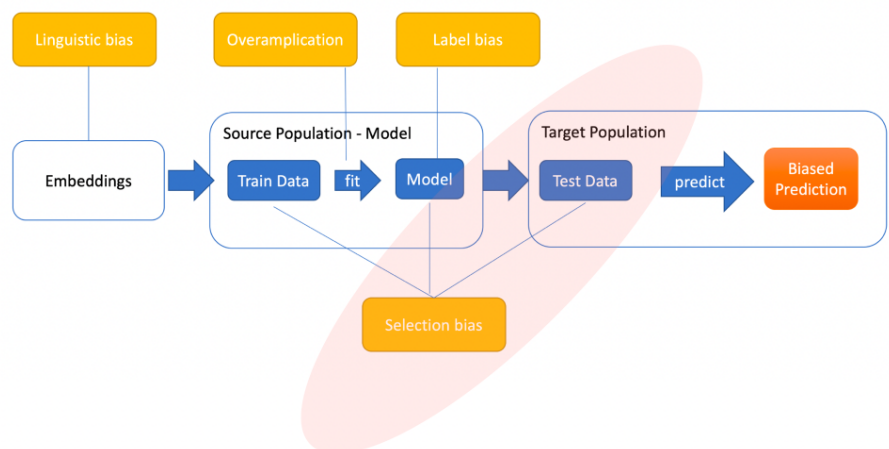

**Figure 1.** Types of predictive bias and their origin. We consider biases of four types[9]: (a) *selection bias* in the training or testing data that are not representative; (b) *label bias* originating from biased human annotations; (c) *textual bias* originating from differences in word distributions; (d) finally, *over-amplification bias* present in statistical models which amplify the discrepancies from the training data. Our framework focuses on the textual bias.

*Annotation bias* is caused by limitations in annotation, i.e., the difficulty of hiring a sufficient number of annotators or those with relevant expertise and the different views and perspectives that caregivers may have. When dealing with text, bias can also be introduced through the language itself (linguistic or *textual bias*). This bias stems from differences in word usage in texts describing different population subgroups. For example, the word "football" may occur more frequently in texts describing males, and the word "cheerleading" may be more often used in association with females. Finally, bias can occur due to *over-amplification* when AI models learn a spurious correlation between the patterns (words) found in the data and the labels instead of relying on relevant information. For example, in the medical domain, chat bots have exhibited subtle sex and race bias when prescribing pain medications[10] or associating women with cooking in semantic role labeling[11].

There has been extensive research in the assessment of bias in AI algorithms in machine learning (ML)[12–15] and natural language processing (NLP)[16–18] as part of broader research on fairness (equitable and non-discriminatory outcomes). However, these general techniques are only beginning to be applied in fields like medicine, and potential remains largely unexplored in the area of mental health[2].

Traditionally, bias mitigation techniques in ML are broadly divided into: pre-processing techniques, learning algorithm modifications, and post-processing techniques. Pre-processing techniques includes approaches like down-weighting the biased instances during training to discourage the model from exaggerating related effects[19,20], as well as resampling methods such as under-sampling or oversampling to balance classes in the data, and data augmentation, which synthesizes samples for underrepresented groups[21,22]. There are also some de-biasing techniques specifically developed in the Natural Language Processing (NLP) domain, such as attribute swapping, which creates new text examples by swapping words indicating sensitive attributes in sentences[23].

Learning algorithm modifications include approaches like adversarial de-biasing which uses adversarial networks to penalize predictions influenced by protected attributes[24,25]. Other approaches also include modifying objective functions to promote fairness during learning (e.g., by penalizing performance differences across demographic groups[26,27]). Recently, self-supervised pre-trained AI models have shown promise in reducing performance disparities across demographic groups[28].

Post-processing techniques include approaches like calibration, which aligns predicted probabilities with actual observed distributions[29]. Popular post-processing approaches transform linguistic embeddings by removing their projections onto the semantic subspace of the demographic aspect[16].

As these methods are applied in the healthcare field, including mental health, several important factors must be taken into account. In healthcare, observed differences between demographic groups can generally be characterized as follows[2]: (a) genuine differences based on biological influences on disease risk (e.g., presence of hormones contributing to breast or prostate cancer), which should be preserved; (b) disparities shaped by non-biological differences (e.g., variations in clinical visit frequency and the writing style of clinical notes resulting from visiting different experts), which should be when possible reduced; and (c) false differences caused by flawed measurement, such as over-diagnosis in certain groups due to misperceptions (e.g., female depression[30]). These errors are primarily annotation flaws that need correction.

This hierarchy of disparities, combined with data sparsity, makes many existing de-biasing methods unsuitable for healthcare. For instance, techniques like swapping gender words between sentences may introduce unrealistic symptom profiles, while removing gender components from word embeddings can distort the meaning of medical terms. Similarly, modifying learning algorithms can lead to overly complex models and poor results when training data is limited.

Additionally, healthcare data is highly heterogeneous, as clinical records often come from various care sites. This heterogeneity remains largely unaddressed by current de-biasing techniques, which typically focus on data from a single source.

In this work, we focus on identifying and mitigating textual bias in demographic subgroups (focusing on sex) found in the text of electronic healthcare records (EHR) in the context of pediatric anxiety—an important issue given the post-pandemic rise in anxiety symptoms among underrepresented groups, particularly females[1].

In the challenging pediatric primary care setting (multiple informants, variety in interpretation of risk factors and symptoms, overlapping symptoms)[3,31,32], AI has the potential to support the healthcare in this challenging setting and under growing pressures. But only if it can offer consistent support across demographic subgroups.

This study aims to detect and mitigate disparities in the textual training data of AI models intended to assist in pediatric mental health screening. These disparities are caused by biological differences and differences in social circumstances (male and female patients with different symptoms were treated in different healthcare sites with different reporting practices) across sex groups and lead to biased predictions.

To address this, we first *assess the presence of bias by evaluating outcome parity across sex subgroups in an AI model trained to predict pediatric anxiety*. Observing disparities in model performance indicates bias, which may disproportionately impact underrepresented groups.

Next, we *identify sources of bias by examining how linguistic and statistical properties in clinical text contribute to unequal outcomes*. Using interpretability techniques, we analyze the impact of gender-related language (e.g., first and last names, gender pronouns) on model predictions to trace the influence of potentially biased terms.

Finally, we *develop and evaluate a data-centric de-biasing method specifically adapted for mental health text*. This method includes normalizing information density to reduce bias and replacing biased words with neutral alternatives. Our approach complements general de-biasing techniques and contributes to best practices in the field, particularly as Large Language Models (LLMs) continue to grow in prominence.

## Methods

Our data come from one of the highest-ranked pediatric institutions in US. We apply a range of state-of-the-art NLP methods to these data.

### Datasets
As part of this study, we created a foundational database consisting of EHR data from the Cincinnati Children's Hospital Medical Center Epic Link. This database has approximately 1.3 million unique patients seen at CCHMC between January 1, 2009, and March 31, 2022, with 63 million clinical notes.

We define anxiety patients as any patients who have ever received any of the diagnostic codes listed in Appendix Table 7. Additional selection criteria are that the patient must have had at least one encounter in the EHR in the 18 months prior to the anxiety diagnosis. As a result, there were 1,383,145 total patients in the CCHMC EHR, 84,426 total anxiety cases that passed our selection criteria, 77,187 total anxiety cases with at least 1 note, 73,288 total anxiety cases with at least 1 note >30 days before their first anxiety diagnosis, and 7,810,849 notes for these 73k patients.

This cohort contains demographic data (sex and race) and the timeline of clinical textual notes provided by different care providers (progress notes, telephone encounters, plan of care notes, patient instructions, etc.).

When we apply our additional selection criteria on note types (we select Progress Notes and Telephone Encounters as deemed the most informative by internal experts), this is reduced further to 4.3 million notes.

*Age binning*

The patients were grouped into age ranges by age (in years) at the time of their anxiety diagnosis. For example, the 5-year-old age group consists of patients who received an anxiety diagnosis between their fifth and sixth birthdays. The controls were matched one-to-one with cases by age and sex. Matched controls 1) were born within 30 days of the case and 2) were of the same sex. Additional criteria for controls were that they had never received one of the anxiety diagnoses at the time of the matched case's anxiety diagnosis and that they had had at least one encounter in the EHR within the 18-month window prior to the case's anxiety diagnosis. Only data prior to the first anxiety diagnosis are used for our analysis.

We used Bins of age 5, 8, 10, 12 and 15 (see the Age Binning subsection below). This selection of bins is not common for pediatrics but gives us a comprehensive selection of datasets with diverse percentages of female patients varying from 36% to 69% (see section Descriptive Analysis in Results)

*Data Cleaning*

To remove duplicates in progress notes and telephone encounters, we first tokenize our notes by removing punctuation and stopwords using the NLTK toolkit[33]. We then vectorized the notes using the CountVectoriser from the Scikit-learn toolkit[34]. Finally, duplicate notes with cosine similarity $\geq 0.8$ were removed. Finally, we selected the 25 most recent notes in each patient's history (average minimum count of notes per patient across the bins). We kept the timelines with at least one record and maintained the 1:1 proportion of cases and controls.

Each final bin contained the training set of ∼3,700–5,064 cases and controls, and the testing set included ∼852–1,278 cases and controls. This study was approved by the Institutional Review Board of Cincinnati Children's Hospital as STUDY2020-0942.

## Anxiety Prediction Models

We built our anxiety prediction models by fine-tuning the state-of-the-art Transformer-based Clinical-BigBird model[35] as imported from HuggingFace[36]. This model is not only pre-trained on clinical text but can also handle long input sequences (up to 4,096 tokens).

We followed the best practices in the domain and fine-tuned the model for 2 epochs. We limited the input length of notes to 1,000 tokens (since considering longer inputs did not result in further improvement). We used AdamW optimizer[37] with a learning rate of 1e-5 and a batch size of 8 (these were the best-performing set of parameters out of the sets suggested for fine-tuning by the model authors[35]).

## Explainability

In addition to measuring the performance of our models with the standard accuracy measure, we perform qualitative analysis of words our models rely on while making predictions. The Local Interpretable Model-Agnostic Explanations (LIME) technique[38] enables this qualitative analysis by offering local explanations and pinpointing specific words that significantly influenced the model decisions. These explanations are "local" because they relate to the model's behavior for each specific incoming note. Note that globally important features (for example, weights that a model assigns to words in its vocabulary) might not be precise enough for the local prediction context.

In our study, we used the LIME methodology to highlight influential words in order to verify if our prediction models are functioning correctly. LIME is designed to uncover undesirable behaviors in AI models that might seem efficient based on standard metrics. For example, Ribeiro et al.[38] showed that a model, despite a 94% accuracy in differentiating documents on Atheism versus Christianity, relied on irrelevant words such as "posting", "host" and "re". These words were wrongly associated with Atheism due to their frequent appearance in the training data.

## Text De-biasing Methods

Motivated by our observations over the textual distributions across demographic subgroups, we propose two following bias mitigation methods:

1. **Information density filtering** (`tf-idf_filt`): we perform normalization of content which involves filtering sentences from concatenated notes using their importance scores. Those importance scores are computed as averaged sum of word-level TF-IDF scores per sentence. TF-IDF scores help identify most salient words in a document by multiplying the frequency of the word in that document by its rarity across all documents (the more rare is the word the higher is the score) so that document keywords (e.g., term "myopia") that may appear multiple times in one document but not in the others receive higher score than auxiliary words such as articles ("the", "a") or verbs ("am", "have").

2. **Gender-word debiasing** (gen_sub): Following the best de-biasing practices from the NLP domain[39], we focus on names and pronouns as gender-biased attributes in text.

    We automatically detect those biased words and replace them with relevant neutral versions. In particular, we detect proper nouns (first and last names) using the off-shelf Stanza tool[40]. We extract unique names and group them based on the character similarity, ensuring that variations of names are considered equivalent (for example, "Jonathan", "Johnathan" and "Johnatan"). This grouping facilitates the creation of a mapping system where each name group is replaced by a generic identifier, such as "person1", "person2", etc. (enumeration is maintained per note).

    Following the name replacement, the text undergoes pronoun substitution, where gender-specific pronouns from a dictionary (for example, "she" and "he") are replaced with their gender-neutral counterparts using a predefined pronoun mapping (for example, "she" and "he" are replaced with "they"). Note that this approach maintains the integrity and coherence of the original text while achieving certain gender neutrality.

Those methods could be used separately or combined. For example, substitution of biased words could be applied to the original text or after the removal of the least informative sentences.

All models used in this study were downloaded and installed locally. All experiments were performed on NVIDIA A100-SXM4-40GB GPUs.

## Results

We elaborate on our findings per each research objective below.

### Objective 1: Assessing the Presence of Bias by Evaluating Outcome Parity Across Sex Subgroups

Best practices in Machine Learning[14,15] measure model bias in terms of classification parity (equality of classification error between groups), anti-classification (effect of protected attributes on predictions) and calibration (difference between predicted risk and factual risk).

In this work with predictive models, we focus on classification parity. Several metrics are commonly used in the fairness literature to evaluate classification parity. These are accuracy equality (equal accuracy between groups), equal opportunity (equal false negative rate (FNR), orientation towards recall), or predictive equality (equal false positive rate (FPR) between groups, orientation towards precision)[41,42]. Following Feldman et al., we also use the balanced error rate (BER), which is the unweighted average of FPR (precision) and FNR (recall):

$$BER = \frac{(\frac{FP}{FP+TN}) + (\frac{FN}{FN+TP})}{2} \tag{1}$$

To assess bias, we compute the ratio of BER scores comparing the non-privileged subgroup to the privileged subgroup. Hence, a ratio of >1 hints towards better performance for the privileged subgroup measured by BER, while a ratio <1 indicates that a model performs better for the non-privileged group. The ratio of >1.25 indicates significant bias towards the privileged subgroup measured by BER, while a ratio of <0.85 indicates that a model performs significantly better for the non-privileged group.

We use FPR, FNR and BER ratio to assess the bias of our five state-of-the-art Transformer-based Clinical-BigBird[35] prediction models built using Bins 5, 8, 10, 12 and 15 with 1K tokens input sequence length (see subsection Anxiety Prediction Models in Methods for more details). Our results are shown in Table 6 across bins and Tables 1 and 2 across demographic subgroups.

Anti-classification measures are also suitable to assess textual bias. For example, predictive models can erroneously rely on biased words (e.g., relying on the pronoun "she" indicating gender to predict the increased probability of depression or on the word "teacher" carrying the semantic of a female in its pre-trained representation[43]). In this work, we use the state-of-the-art interpretability analysis to verify if our models erroneously rely on gender words (first and last names, as well as gender pronouns). Note that this reliance could also be considered the sub-case of over-amplification bias (picking up on imperfect evidence to predict outcomes) but while working with text we consider it as textual bias. We leave the investigation of calibration measures to future work.

Overall performance of our models is reported in Table 6. Our model achieves the performance of 0.61 accuracy on average across test data (this performance is above the random 0.50 accuracy). We also report the percentage of uncertain predictions as an indicator of the model's confidence in its decisions. Standard AI predictive models issue a probability distribution over their outcomes. If the predicted probability is ≥0.5, the instance is classified as the positive class. Uncertain predictions (from the borderline probability zone [0.4, 0.6]) are less stable and reliable. If the percentage of such predictions is high a model can hardly be considered useful.

Looking closer at the bias within demographic subgroups using BER, we observe that our Bin 10 model is slightly more biased towards male patients with the BER ratio of 1.33 (see Table 1). BER ratios however do not tell us the whole story. Looking at FNR scores (equal opportunity), we can see that female patients get consistently under-diagnosed (+0.09 FNR average higher for females than for males; an increase of +0.13 is observed for bins 5 and 15). This trend is observed across cohorts and is consistent even if the percentage of female patients changes. FPR values do not exhibit such a systematic increase for males or females and stays roughly the same. Also, the percentage of uncertain predictions is systematically higher for females than for males (+5%) confirming that our models "hesitate" more for female patients. Finally, accuracy differences exhibit a decrease of -0.04 point on average for female patients.

In summary, FNR (equal opportunity) is the most expressive measure revealing relevant predictive bias in our anxiety models. It correlates well with differences in textual statistics from the Descriptive Analysis subsection. Bias is most pronounced for female patients in cohorts 5 and 15, where the similarities in textual distributions are the lowest.

For the anti-classification analysis (reliance on biased words for prediction), we check the top ten examples with highest model confidence (most useful predictions) in a case or a control prediction (20 examples per age bin, 100 examples in total across bins). We extract five words per example that influence the model decision the most and then collate them into a frequency vocabulary per predicted class. The results of this analysis demonstrate that the models mostly rely on relevant words both for cases and controls across low and high confidence groups (Table 3). However, there is some tendency to rely on irrelevant gender-biased words: for example, for cases 10% from the collated dictionary is biased.

Note that similar tendencies are observed across race subgroups: we observe an average increase in FNR of +0.05 for the non-privileged race group. FPR values exhibit an increase towards the privileged class (+0.05 FPR). Also, a decrease of -0.03 in accuracy is observed for other races.

**Objective 2: Identifying Bias Sources Through Textual Distribution Analysis**

To understand the reasons for performance differences, we analyzed primarily the sex subgroups of the dataset and the properties of the relevant texts. In particular, we considered the following characteristics of textual distributions commonly applied in the NLP domain:

1. *Average length* of a patient note in words (without tokenization, i.e., separation of punctuation). We measured the concatenation of all the valid notes per patient.

2. *Percentage of medical terms*, in each note on average. We extract biomedical named entities using the state-of-the-art Stanza tool[44]. Those in-domain entities are extracted for ten standard categories such as Observation, Treatment, Anatomy, Procedure, etc.

3. *Percentage of gender-biased words*, in each patient note on average. Using the best practices from NLP[39], we extract all the proper nouns (first and last names) using the off-shelf tool[40], as well as gender pronouns (he, she, his, her, him, hers). We focus on this semantic group as the most relevant to our cohort design involving matching by age and sex.

4. *Jaccard distance* measures the similarity between two vocabularies by estimating the portion of common words. It is computed by dividing the number of words shared by both vocabularies by the total number of words in both vocabularies combined. We mainly compare vocabularies for male/female.

5. *Familiarity score* assesses the ratio of common words to unique words. It is computed dividing the proportion of words which occur in both sets of words by the uniqueness ratio, the ratio of words that occur only in one of the sets. Again we mainly compare vocabularies male/female.

The results of our analysis across bins are presented in Table 4 and across demographic groups within each bin in Table 5.

Our initial observations indicate notable changes in demographics and text properties across age bands (seen in Table 4). The ratio of females is consistently growing (from 36% for Bin 5 to 69% for Bin 15), while the percentage of the other races is stable (around 30%). The average concatenated note length remains constant at around 4K tokens.

The differences between demographic groups within the bins are clear (see Tables 5. The percentage of cases is always 50% across sex subgroups, which is predefined by the cohort design. This percentage is typically lower for the non-privileged race subgroup as race was not factored into the cohort design (43% on average for other races). There are also obvious differences in the length of the notes. Notes for the male subgroup are on average ~540 words longer (~180 words longer for white patients).

|   |   | Male |   |   |   |   | Female |   |   |   |   |   |
|---|---|---|---|---|---|---|---|---|---|---|---|---|
|   |   | Acc | unc | FPR | FNR | BER | Acc | unc | FPR | FNR | BER | BER r. |
| Bin 5 | Orig | 0.58 | 22 | 0.61 | 0.24 | 0.42 | 0.56 | 31 | 0.52 | 0.37 | 0.44 | 1.04 |
|  | rnd_filt | 0.61 | 24 | 0.54 | 0.25 | 0.39 | 0.57 | 33 | 0.47 | 0.39 | 0.43 | 1.09 |
|  | tf-idf_filt | 0.56 | 18 | 0.55 | **0.34** | 0.44 | 0.58 | 22 | 0.49 | **0.36** | 0.42 | 0.95 |
|  | gen_sub | 0.59 | 27 | 0.42 | 0.41 | 0.41 | 0.61 | 29 | 0.29 | 0.48 | 0.39 | 0.95 |
|  | tf-idf_filt+gen_sub | 0.59 | 11 | 0.34 | 0.47 | 0.41 | 0.59 | 9 | 0.31 | 0.51 | 0.41 | 1.00 |
| Bin 8 | Orig | 0.64 | 21 | 0.14 | 0.58 | 0.36 | 0.60 | 25 | 0.17 | 0.62 | 0.40 | 1.11 |
|  | rnd_filt | 0.66 | 14 | 0.17 | 0.51 | 0.34 | 0.62 | 17 | 0.17 | 0.59 | 0.38 | 1.11 |
|  | tf-idf_filt | 0.57 | 13 | 0.44 | **0.42** | 0.43 | 0.61 | 11 | 0.35 | **0.42** | 0.39 | 0.90 |
|  | gen_sub | 0.68 | 12 | 0.13 | 0.52 | 0.32 | 0.60 | 14 | 0.16 | 0.64 | 0.40 | 1.24 |
|  | tf-idf_filt+gen_sub | 0.64 | 12 | 0.31 | 0.42 | 0.36 | 0.63 | 12 | 0.22 | 0.52 | 0.37 | 1.02 |
| Bin 10 | Orig | 0.67 | 29 | 0.37 | 0.28 | 0.33 | 0.57 | 42 | 0.51 | 0.36 | 0.43 | 1.33 |
|  | rnd_filt | 0.65 | 39 | 0.42 | 0.29 | 0.35 | 0.61 | 39 | 0.41 | 0.37 | 0.39 | 1.11 |
|  | tf-idf_filt | 0.66 | 31 | 0.41 | 0.27 | 0.34 | 0.58 | 35 | 0.43 | 0.41 | 0.42 | 1.24 |
|  | gen_sub | 0.62 | 31 | 0.45 | 0.31 | 0.38 | 0.59 | 36 | 0.42 | 0.40 | 0.41 | 1.08 |
|  | tf-idf_filt+gen_sub | 0.60 | 22 | 0.26 | **0.53** | 0.40 | 0.61 | 17 | 0.19 | **0.58** | 0.39 | 0.98 |
| Bin 12 | Orig | 0.62 | 14 | 0.48 | 0.28 | 0.38 | 0.64 | 20 | 0.36 | 0.36 | 0.36 | 0.95 |
|  | rnd_filt | 0.64 | 14 | 0.28 | 0.44 | 0.36 | 0.62 | 13 | 0.23 | 0.53 | 0.38 | 1.06 |
|  | tf-idf_filt | 0.62 | 8 | 0.37 | 0.39 | 0.38 | 0.63 | 11 | 0.26 | 0.48 | 0.37 | 0.97 |
|  | gen_sub | 0.63 | 12 | 0.20 | 0.54 | 0.37 | 0.61 | 12 | 0.10 | 0.68 | 0.39 | 1.06 |
|  | tf-idf_filt+gen_sub | 0.61 | 18 | 0.48 | **0.31** | 0.39 | 0.61 | 15 | 0.39 | **0.39** | 0.39 | 0.99 |
| Bin 15 | Orig | 0.66 | 35 | 0.26 | 0.42 | 0.34 | 0.61 | 29 | 0.24 | 0.55 | 0.39 | 1.15 |
|  | rnd_filt | 0.65 | 16 | 0.29 | 0.42 | 0.35 | 0.59 | 17 | 0.29 | 0.52 | 0.41 | 1.15 |
|  | tf-idf_filt | 0.60 | 16 | 0.44 | **0.36** | 0.40 | 0.61 | 18 | 0.35 | **0.43** | 0.39 | 0.97 |
|  | gen_sub | 0.64 | 15 | 0.33 | 0.38 | 0.36 | 0.58 | 15 | 0.35 | 0.48 | 0.42 | 1.16 |
|  | tf-idf_filt+gen_sub | 0.62 | 18 | 0.40 | 0.36 | 0.38 | 0.58 | 19 | 0.41 | 0.44 | 0.42 | 1.12 |

**Table 1.** Classification parity results across sex groups for models trained with original and de-biased data. We report accuracy, percentage of uncertain predictions (probability in [0.4, 0.6]), False Positive Rate (FPR), False Negative Rate (FNR) and Balanced Error Rates (BER). Red highlights higher values of FNR for non-privileged group. Green highlights reduction in FNR to benefit the non-privileged group. Bold highlights best results. We also highlight the BER ratio indicating bias and its improvement.

|   |   | White |   |   |   |   | Other |   |   |   |   |   |
|---|---|---|---|---|---|---|---|---|---|---|---|---|
|   |   | Acc | unc | FPR | FNR | BER | Acc | unc | FPR | FNR | BER | BER r. |
| Bin 5 | Orig | 0.55 | 25 | 0.59 | 0.31 | 0.45 | 0.55 | 30 | 0.56 | 0.35 | 0.45 | 1.01 |
|  | tf-idf_filt+gen_sub | 0.59 | 10 | 0.36 | 0.47 | 0.41 | 0.59 | 11 | 0.33 | 0.50 | 0.41 | 1.00 |
| Bin 8 | Orig | 0.64 | 24 | 0.17 | 0.56 | 0.36 | 0.62 | 19 | 0.13 | 0.62 | 0.38 | 1.04 |
|  | tf-idf_filt+gen_sub | 0.65 | 12 | 0.27 | 0.43 | 0.35 | 0.64 | 13 | 0.25 | 0.47 | 0.36 | 0.96 |
| Bin 10 | Orig | 0.52 | 37 | 0.54 | 0.42 | 0.48 | 0.65 | 34 | 0.38 | 0.33 | 0.35 | 0.74 |
|  | tf-idf_filt+gen_sub | 0.59 | 20 | 0.27 | 0.56 | 0.41 | 0.61 | 21 | 0.23 | 0.56 | 0.39 | 0.95 |
| Bin 12 | Orig | 0.63 | 20 | 0.45 | 0.29 | 0.37 | 0.64 | 14 | 0.36 | 0.36 | 0.35 | 0.95 |
|  | tf-idf_filt+gen_sub | 0.63 | 14 | 0.49 | 0.25 | 0.37 | 0.62 | 14 | 0.39 | 0.37 | 0.38 | 1.03 |
| Bin 15 | Orig | 0.66 | 36 | 0.26 | 0.42 | 0.34 | 0.59 | 32 | 0.22 | 0.60 | 0.41 | 1.22 |
|  | tf-idf_filt+gen_sub | 0.63 | 17 | 0.34 | 0.40 | 0.37 | 0.57 | 22 | 0.37 | 0.50 | 0.43 | 1.17 |

**Table 2.** Classification parity results across race groups for models trained with original and de-biased data. We report accuracy, percentage of uncertain predictions (probability in [0.4, 0.6]), False Positive Rate (FPR), False Negative Rate (FNR) and Balanced Error Rates (BER). Red highlights higher values of FNR for non-privileged group. Green highlights reduction in FNR to benefit the non-privileged group. We also highlight the BER ratio indicating bias and its improvement.

|  | Top-10 Av. Influential | Top-10 Av. Frequency | biased,% | biased, av fr |
| --- | --- | --- | --- | --- |
| | Original | | | |
| Case | cognitive, of, learning, therapist, attention, educational, psychotherapy, she, psychology, skills | anxiety, and, depression, mood, disorder, impulsivity, to, coping, depressed, behavior | 10 | 1 |
| Ctrl | play, disorder, poison, accepted, letters, on, homework, O, form, pt | eye, the, glasses, hearing, vision, ear, audiologic, eyes, amblyopia, negative | 4 | 1.2 |
| | `tf-idf_filt` | | | |
| Case | amblyopia, allergies, asthma, seizure, eczema, reassurance, psoriasis, deficit, presents, stress | anxiety, disorder, depression, message, contact, mood, and, to, suicidal, from | 3 | 4.5 |
| Ctrl | reassurance, seasonal, headaches, reports, diagnosis, myopia, pain, including, cycloplegic, FHx | eye, hearing, ear, the, with, complaint, exam, chief, presents, vision | 2 | 1 |
| | `tf-idf_filt+gen_sub` | | | |
| Case | seizure, seizures, observation, respiratory, presents, developmental, distractibility, anxieties, contact, suicidal | anxiety, contact, depression, message, the, AM, EDT, and, behavior, suicidal | 6 | 2.14 |
| Ctrl | presents, with, up, [formatting sign], elbow, pain, discharge, diagnosis, struggling, language | eye, the, exam, ear, hearing, with, complaint, chief, no, vision | 3 | 1 |

**Table 3.** Top-10 most influential words and Top-10 most frequent words as extracted by the state-of-the-art interpretability LIME tool[38] from 100 examples across bins 5, 8, 10, 12 and 15. We analyze ten examples from each set of predictions with the highest confidence in case or control outcomes (20 examples per bin, 100 examples in total). We extract five most important word per example and collate them into a frequency vocabulary of unique words per confidence-outcome group. We average their importance scores. We also report the percentage gender-biased words (first and last names, as well as gender pronouns: he, she, his, her, him, hers) we observe in each vocabulary, as well as their average frequency.

The word distributions for male/female show the similarity of 0.54 Jaccard on average (average Familiarity 2.4). The lowest similarity for male/female is observed for Bins 5 and 15 (average Jaccard 0.43, Familiarity index 1.75). The percentages of terms and biased words for male/female are roughly the same. The distributions of terms male/female are even more dissimilar (average Jaccard 0.34, average Familiarity 1.5). Relatively low similarity values for word distributions are also observed across race subgroups (average Jaccard index 0.34, average Familiarity index 1.52). Though seen that other races make only 30% of the patients those values are not directly comparable to the ones for sex subgroups.

Given that the female subgroup is well-represented (49% of the examples on average), the differences we observe can not be explained by scarcity and suggest qualitative differences in the content of the notes across demographic subgroups (as evidenced by rather low similarity scores). The volume of diagnostic and gender-biased content remains constant across sex groups. This suggests that the differences in the length of the notes across subgroups are caused by other than diagnostic content and may be filtered. These content differences may be caused by the fact that notes of female/male patients tend to come from different care sites following different reporting standards. For example, the male notes in bin 5 come from more than 400 care sites, while the female notes come from 337 sites. The female notes come mostly from General, Developmental and Behavioral Pediatrics, while male notes very often come from more specialised departments such as Neurology and Gastroenterology.

In summary, our cohort matching procedure considering age/sex ensures we have equivalent representation of each between cases/controls. However, statistical differences in textual distributions for sex subgroups persist. It is important to note that these differences are influenced by symptomatic variations and social circumstances of patients of different sexes who are seen at different care sites. These differences are difficult to control (it is difficult to control lengths of notes across care sites or choose only the records without gender-related words or names), but could be more easily manipulated with the help of text pre-processing techniques. We propose two techniques like this for bias mitigation in subsection Bias Mitigation below.

**Objective 3: Developing and Evaluating a De-biasing Method for Mental Health Text Data**

We have already seen that there are considerable differences in information density and textual distributions of notes for sex subgroups, in particular, those differences naturally involve gender words which our models can erroneously rely on. Our core hypothesis is that, by eliminating less relevant sentences (`tf-idf_filt`) we can balance density in the notes and reduce bias in our models. Regarding gender words, we have already seen that male and female notes contain similar percentages of such words. Hence we have developed an approach to neutralize those words via substitution with their gender-neural versions rather

|                      | Bin 5 | Bin 8 | Bin 10 | Bin 12 | Bin 15 |
|----------------------|-------|-------|--------|--------|--------|
| count, total         | 4188  | 3884  | 3656   | 3662   | 5064   |
| %, cases             |       |       | 50     |        |        |
| %, Male              | 64    | 60    | 56     | 45     | 31     |
| %, Female            | 36    | 40    | 44     | 55     | 69     |
| %, White             | 68    | 72    | 74     | 74     | 76     |
| %, Other             | 32    | 28    | 26     | 26     | 24     |
| Av length            | 3887  | 4162  | 3957   | 3783   | 3916   |
| Av terms, %          | 21    | 20    | 20     | 21     | 21     |
| Av biased words, %   | 3.0   | 3.1   | 3.2    | 3.2    | 3.0    |

**Table 4.** Demographic statistics and properties of textual notes across Age bins (measured for the training data). We report average values per concatenated patient note.

|                                | Bin 5 |      | Bin 8 |      | Bin 10 |      | Bin 12 |      | Bin 15 |      |
|--------------------------------|-------|------|-------|------|--------|------|--------|------|--------|------|
|                                | M     | F    | M     | F    | M      | F    | M      | F    | M      | F    |
| %, cases                       |       |      |       |      | 50     |      |        |      |        |      |
| Av length                      | 4139  | 3443 | 4465  | 3710 | 4281   | 3549 | 4061   | 3556 | 3935   | 3908 |
| Av model length                | 729   | 715  | 735   | 731  | 733    | 718  | 732    | 726  | 731    | 737  |
| Jaccard ind (↑), all vocab     | 0.44  |      | 0.51  |      | 0.59   |      | 0.74   |      | 0.42   |      |
| Familiarity ind (↑), all vocab | 1.77  |      | 2.04  |      | 2.43   |      | 3.83   |      | 1.72   |      |
| Av terms, %                    | 20    | 21   | 20    | 21   | 20     | 21   | 20     | 21   | 20     | 21   |
| Jaccard ind (↑), terms         | 0.51  |      | 0.59  |      | 0.67   |      | 0.68   |      | 0.41   |      |
| Familiarity ind (↑), terms     | 2.02  |      | 2.45  |      | 3.05   |      | 3.17   |      | 1.69   |      |
| Av biased words, %             | 3.0   | 2.9  | 3.2   | 3.1  | 3.2    | 3.2  | 3.2    | 2.9  | 3.2    | 3.0  |
|                                | W     | O    | W     | O    | W      | O    | W      | O    | W      | O    |
| %, cases                       | 51    | 47   | 53    | 43   | 53     | 42   | 53     | 41   | 53     | 41   |
| Av length                      | 4021  | 3605 | 4215  | 4022 | 4026   | 3756 | 3788   | 3770 | 3893   | 3987 |
| Jaccard ind (↑), all vocab     | 0.39  |      | 0.34  |      | 0.31   |      | 0.33   |      | 0.31   |      |
| Familiarity ind (↑), all vocab | 1.65  |      | 1.53  |      | 1.44   |      | 1.50   |      | 1.46   |      |
| Terms, %                       | 21    | 22   | 20    | 20   | 20     | 21   | 20     | 21   | 21     | 21   |
| Jaccard ind (↑), terms         | 0.43  |      | 0.36  |      | 0.33   |      | 0.34   |      | 0.30   |      |
| Familiarity ind (↑), terms     | 1.77  |      | 1.55  |      | 1.48   |      | 1.50   |      | 1.44   |      |
| Av gender-biased words, %      | 3.1   | 2.7  | 3.2   | 2.9  | 3.2    | 3.0  | 3.3    | 3.1  | 3.1    | 2.9  |

**Table 5.** Class statistics and properties of textual notes across sex and race demographic subgroups in Bins. We report average values per concatenated patient note. M indicates Males, F indicates Females, W and O indicate White and Other races respectively. Distributional similarities (Jaccard and Familiarity indices) are reported between demographic subgroups within each bin.

|                       | Bin 5 |     | Bin 8 |     | Bin 10 |     | Bin 12 |     | Bin 15 |     |
|-----------------------|-------|-----|-------|-----|--------|-----|--------|-----|--------|-----|
|                       | Acc   | unc | Acc   | unc | Acc    | unc | Acc    | unc | Acc    | unc |
| Original              | 0.58  | 27  | 0.62  | 22  | 0.60   | 35  | 0.63   | 17  | 0.63   | 32  |
| `rnd_filt`            | 0.58  | 28  | 0.63  | 15  | 0.64   | 38  | 0.62   | 14  | 0.62   | 17  |
| `tf-idf_filt`         | 0.57  | 20  | 0.60  | 12  | 0.62   | 32  | 0.62   | 10  | 0.60   | 18  |
| `gen_sub`             | 0.60  | 27  | 0.63  | 13  | 0.61   | 33  | 0.62   | 12  | 0.62   | 15  |
| `tf-idf_filt+gen_sub` | 0.59  | 10  | 0.63  | 12  | 0.62   | 19  | 0.61   | 16  | 0.60   | 18  |

**Table 6.** Classification parity results for models trained on original and de-biased notes. We report accuracy and the percentage of uncertain predictions (unc, percentage of model predictions with probabilities from the borderline probability zone [0.4, 0.6].

than their removal (`gen_sub`, see subsection Text De-biasing Methods in Methodology).

We applied our de-biasing methods as described in the Methodology section above to modify our training data. In particular, we applied the `rnd_filt` method and removing 20% of sentences at random as our baseline. We compared it to the performance of our approach `tf-idf_filt` which removes 20% sentences according to their informativeness score (threshold defined empirically). We applied the `gen_sub` method on the original notes, as well as on the notes filtered with `tf-idf_filt` to trace the effect of both models combined. We compare our approach to the baseline approach where we remove 20% of sentences chosen at random (`rnd_filt`).

Each time we modified the training data, we re-trained our models to obtain new models, which were then tested on the original test data.

Our results show that the de-biased models in general maintain the performance of the original models (Table 6): both `gen_sub` and `rnd_filt` slightly increase the performance by +0.5 accuracy, while `tf-idf_filt` decreases the performance on average by -1 accuracy. The mixed method `tf-idf_filt+gen_sub` maintains the original performance with negligible changes. While maintaining performance, our de-biasing methods exhibit positive influence on the performance by reducing the percentage of uncertain predictions by -8% on average with the highest reduction for the `tf-idf_filt+gen_sub` of -12 %.

In terms of the reduction of bias as measured by FNR, `tf-idf_filt` is a clear winner (see Table 1). It outperforms the random sentence removal baseline `rnd_filt`, reducing the FNR gap by -0.024 (27%, initial average gap 0.09) point on average with the highest reduction by -0.11 point for bin 5 (from 0.13 to 0.02) and -0.06 point for bin 15 (from 0.13 to 0.07). `rnd_filt` baseline does not exhibit any consistent behavior and does not influence the gap across bins. Our `gen_sub` approach when applied alone is not efficient to reduce the FNR gap as well. On average we even observe a small increase in this gap rather than a decrease (+0.008). `tf-idf_filt+gen_sub` has roughly the same performance as `tf-idf_filt` with the average decrease of -0.022 point. Both `tf-idf_filt` and `tf-idf_filt+gen_sub` approaches manage to maintain the BER ratios within the acceptable level, and even decrease the BER ratio for Bin 10 from 1.33 to the acceptable values of 1.24 and 0.98 for `tf-idf_filt` and `tf-idf_filt+gen_sub` respectively. `tf-idf_filt` reduces the increase in uncertain predictions for females by 50%, while `tf-idf_filt+gen_sub` fully eliminates this dis-balance.

`tf-idf_filt+gen_sub` has also capacity to mitigate bias for race subgroups reducing the FNR gap by -0.034 point on average. For bin 10, we also reduce the bias towards the non-privileged class (from BER ratio 0.74 to the acceptable level of 0.95).

We also observe a positive effect of our de-biasing methods in terms of the words our classifiers rely on (anti-classification, see Table 3)). Our first observation is that both `tf-idf_filt` and `tf-idf_filt+gen_sub` improve the generalizability of our models. We observe less terms appear in the statistics signifying more reliance on context (e.g., words "complaint", "presents", "no") while making predictions rather than overfitting to keywords (e.g., "anxiety", "depression", etc.). Furthermore, our de-biasing techniques reduce the percentage of biased words our models rely on: for example, `tf-idf_filt` reduces this percentage from 10% to 3% for cases, from 4% to 2% for controls. `tf-idf_filt+gen_sub` further reduces twice the frequency of biased words `tf-idf_filt` relies on. This is a positive effect since those residual biased associations become less systematic as compared to `tf-idf_filt`.

## Discussion

This study aimed to detect and mitigate linguistic bias in training data of AI models intended to assist in pediatric mental health screening, focusing on sex bias. First, we found measurable disparities in AI model performance across sex subgroups, highlighting predictive bias that disproportionately affects females. This was evident in the lower classification parity (4% lower accuracy for females than for males across age groups) and higher false negative rates for female patients (9% higher on average across age groups), suggesting that the model was less accurate in diagnosing anxiety in this subgroup. Second, we identified intrinsic differences in textual properties between male and female patient notes, such as variations in note length (notes for males are 500 words longer), word distribution (low similarities for male/female word distributions of 0.54 Jaccard index, whereas values above 0.7 are considered indicative), and information density (low similarities for male/female term distributions of 0.34 Jaccard index). These differences are likely linked to reporting practices and medical documentation styles, which contribute to biased outcomes in AI predictions. Third, our data-centric approach to mitigate this bias using information density filtering and gender-neutral word substitutions improved classification parity by up to 27%, particularly benefiting the non-privileged subgroup (females).

This study supports previous findings that AI models trained on clinical data can perpetuate biases present in the original data, disproportionately affecting underrepresented groups, e.g., the study of Obermeyer et al. has found that commercial prediction algorithms used in healthcare to identify patients with complex needs exhibit significant racial bias, as they predict healthcare costs instead of illness severity, resulting in Black patients being under-identified for additional care despite having more severe health conditions[45]. Similarly, our findings align with prior research indicating that linguistic patterns, such as gendered language, contribute to bias in natural language processing models used in healthcare[18,46].

While previous studies have shown that AI models can produce biased outcomes across demographic groups, addressing bias in healthcare data presents unique challenges. In healthcare, it is important to retain biological differences in the training data that reflect actual patient needs, while mitigating biases that arise from non-biological factors, such as cultural or provider-based documentation differences. Traditional NLP methods, such as swapping gendered words[23] or removing gendered meanings from word representations[16], are not suitable for this purpose in healthcare, as they could lead to inaccurate data representations. Moreover, healthcare notes vary widely across providers, introducing additional complexity. Our method addresses these challenges by selectively de-biasing data: it maintains information relevant to clinical care while reducing the influence of biased language and normalizing information density across records.

This study has several strengths. First, this study focuses on a data-centric approach, emphasizing the quality and relevance of data rather than improving algorithms (model-centric AI)[47,48]. Second, the de-biasing methodology developed here is specifically adapted for heterogeneous healthcare text data from different clinical sites. This tailored approach is particularly effective in pediatric mental health, where reliable and equitable early detection is critical. This approach not only clarifies how specific language influences model predictions but also demonstrates practical effectiveness: bias mitigation techniques, such as word substitution and information density balancing, reduced diagnostic bias by up to one-third for systematically under-diagnosed female patients. Third, this study creates a pathway to further exploration of complimentary de-biasing techniques specific to AI in the mental health domain.

The study has some limitations. The focus of the study relies on the quality and consistency of the EHR text. Variability in note quality between providers may influence the model's ability to generalize across different clinical settings, which is a common challenge in EHR research. Also, the study focuses on only one type of demographic bias within a male/female pediatric population.

## Conclusion

Anxiety disorders are a leading cause of disability in children and adolescents worldwide, with rising rates among minority groups. AI can play a transformative role in early mental health detection, but its success depends on reliable, unbiased data. This study presents a data-centric de-biasing approach designed to address disparities in AI model performance in clinical text, especially among under-diagnosed groups like female patients. By balancing information density and neutralizing biased terms, our approach reduced diagnostic bias by up to one-third. These findings underscore the importance of bias-aware data processing to create fair and effective AI tools in mental health.

## Code and Data Availability

The code and data could not be publicly shared due to their confidentiality requirement. The code was implemented in Python 3.10.

## Acknowledgments

This work was funded by Cincinnati Children's Hospital Medical Center's Mental Health Trajectory program. The views expressed are those of the authors and not necessarily those of the Cincinnati Children's Hospital Medical Center's Decode program. This work was authored in part by UT-Battelle, LLC, under Contract No. DE-AC05-00OR22725 with the U.S. Department of Energy. The United States Government retains and the publisher, by accepting the article for publication, acknowledges that the United States Government retains a non-exclusive, paid-up, irrevocable, world-wide license to publish or reproduce the published form of this manuscript or allow others to do so, for United States Government purposes. The Department of Energy will provide public access to these results of federally sponsored research in accordance with the DOE Public Access Plan (http://energy.gov/downloads/doe-public-access-plan).

## Author contributions statement

JI conceived the approach and experiments. DS prepared the datasets. JI processed the datasets, conducted the experiments, analyzed the results, and wrote the paper. PB assisted in writing the manuscript. VY assisted in conducting experiments. TC assisted in the conceptualization. JS, GA, JT, SC, MC, and AK participated in the project and assisted with manuscript editing. All reviewed the research and the manuscript. All authors approved the manuscript.

## Additional information

**Competing interests**: The work is in collaboration with Cincinnati Children's Hospital Medical Center University College London and Queen Mary University of London.

# A  Anxiety Diagnosis codes

| Vocabulary | Code | Description |
| --- | --- | --- |
| ICD9CM | 293.84 | Anxiety disorder in conditions classified elsewhere |
| ICD9CM | 300 | Anxiety, dissociative and somatoform disorders |
| ICD9CM | 300.00 | Anxiety state, unspecified |
| ICD9CM | 300.01 | Panic disorder without agoraphobia |
| ICD9CM | 300.02 | Generalized anxiety disorder |
| ICD9CM | 300.09 | Other anxiety states |
| ICD9CM | 300.2 | Phobic disorders |
| ICD9CM | 300.21 | Agoraphobia with panic disorder |
| ICD9CM | 300.22 | Agoraphobia without mention of panic attacks |
| ICD9CM | 300.23 | Social phobia |
| ICD9CM | 300.29 | Other isolated or specific phobias |
| ICD9CM | 309.21 | Separation anxiety disorder |
| ICD9CM | 309.24 | Adjustment disorder with anxiety |
| ICD9CM | 309.28 | Adjustment disorder with mixed anxiety and depressed mood |
| ICD9CM | 313 | Disturbance of emotions specific to childhood and adolescence |
| ICD10CM | F06.4 | Anxiety disorder due to known physiological condition |
| ICD10CM | F12.280 | Cannabis dependence with cannabis-induced anxiety disorder |
| ICD10CM | F12.980 | Cannabis use, unspecified with anxiety disorder |
| ICD10CM | F13.980 | Sedative, hypnotic or anxiolytic use, unspecified with sedative, hypnotic or anxiolytic-induced anxiety disorder |
| ICD10CM | F15.280 | Other stimulant dependence with stimulant-induced anxiety disorder |
| ICD10CM | F16.980 | Hallucinogen use, unspecified with hallucinogen-induced anxiety disorder |
| ICD10CM | F19.280 | Other psychoactive substance dependence with psychoactive substance-induced anxiety disorder |
| ICD10CM | F19.980 | Other psychoactive substance use, unspecified with psychoactive substance-induced anxiety disorder |
| ICD10CM | F40.240 | Claustrophobia |
| ICD10CM | F41.0 | Panic disorder [episodic paroxysmal anxiety] |
| ICD10CM | F41.1 | Generalized anxiety disorder |
| ICD10CM | F41.3 | Other mixed anxiety disorders |
| ICD10CM | F41.8 | Other specified anxiety disorders |
| ICD10CM | F41.9 | Anxiety disorder, unspecified |
| ICD10CM | F43.22 | Adjustment disorder with anxiety |
| ICD10CM | F43.23 | Adjustment disorder with mixed anxiety and depressed mood |
| ICD10CM | F60.6 | Avoidant personality disorder |

**Table 7.** The list of diagnosis codes used to define an anxiety diagnosis for this analysis. Cincinnati Children's Hospital Medical Center switched from the International Classification of Diseases (ICD) version 9 to version 10 on October 1, 2015.